\documentclass[lettersize,journal]{IEEEtran}
\usepackage{amsmath,amsfonts}
\usepackage{algorithmic}
\usepackage{algorithm}
\usepackage{array}
\usepackage[caption=false,font=normalsize,labelfont=sf,textfont=sf]{subfig}
\usepackage{textcomp}
\usepackage{stfloats}
\usepackage{url}
\usepackage{verbatim}
\usepackage{graphicx}
\usepackage{cite}
\usepackage{color}
\usepackage{booktabs}
\begin{document}

\title{HiT: Building Mapping with Hierarchical Transformers}

\author{Mingming Zhang, Qingjie Liu,~\IEEEmembership{Member,~IEEE,} and Yunhong Wang,~\IEEEmembership{Fellow,~IEEE}
\thanks{Mingming Zhang, Qingjie Liu, and Yunhong Wang are with the State Key Laboratory of Virtual Reality Technology and Systems, Beihang University, Beijing 100191, China, and also with the Hangzhou Innovation Institute, Beihang University, Hangzhou 310051, China (e-mail: sara\_@buaa.edu.cn; qingjie.liu@buaa.edu.cn; yhwang@buaa.edu.cn). This work was supported by Science and Technology Innovation 2030-Key Project of ``New Generation Artificial Intelligence'' under Grant 2020AAA0108205 and NSFC No. 62176017.}
}

\markboth{IEEE TRANSACTIONS ON GEOSCIENCE AND REMOTE SENSING}%
{Shell \MakeLowercase{\textit{et al.}}: A Sample Article Using IEEEtran.cls for IEEE Journals}


\maketitle

\begin{abstract}
Deep learning-based methods have been extensively explored for automatic building mapping from high-resolution remote sensing images over recent years. While most building mapping models produce vector polygons of buildings for geographic and mapping systems, dominant methods typically decompose polygonal building extraction in some sub-problems, including segmentation, polygonization, and regularization, leading to complex inference procedures, low accuracy, and poor generalization. In this paper, we propose a simple and novel building mapping method with Hierarchical Transformers, called HiT, improving polygonal building mapping quality from high-resolution remote sensing images. HiT builds on a two-stage detection architecture by adding a polygon head parallel to classification and bounding box regression heads. HiT simultaneously outputs building bounding boxes and vector polygons, which is fully end-to-end trainable. The polygon head formulates a building polygon as serialized vertices with the bidirectional characteristic, a simple and elegant polygon representation avoiding the start or end vertex hypothesis. Under this new perspective, the polygon head adopts a transformer encoder-decoder architecture to predict serialized vertices supervised by the designed bidirectional polygon loss. Furthermore, a hierarchical attention mechanism combined with convolution operation is introduced in the encoder of the polygon head, providing more geometric structures of building polygons at vertex and edge levels. Comprehensive experiments on two benchmarks (the CrowdAI and Inria datasets) demonstrate that our method achieves a new state-of-the-art in terms of instance segmentation and polygonal metrics compared with state-of-the-art methods. Moreover, qualitative results verify the superiority and effectiveness of our model under complex scenes.
\end{abstract}

\begin{IEEEkeywords}
Building mapping, transformer, bidirectional polygon loss, hierarchical attention mechanism.
\end{IEEEkeywords}

\section{Introduction}
\IEEEPARstart{B}{uilding} mapping from remote sensing images is an essential task for geographic and mapping applications, including disaster management and assessment, city planning, human activity monitoring, and demographics. Deep learning methods have emerged over recent years due to their powerful representation learning and success in many tasks (\textit{e.g.}, classification, detection, segmentation). Meanwhile, the quick development of satellite and sensor techniques makes large-scale high-resolution remote sensing images easy to access, and some building segmentation benchmarks have been built for automatic building extraction. Therefore, deep learning-based building mapping from high-resolution remote sensing images has attracted more and more attention in the remote sensing community.

Early deep learning-based methods apply semantic segmentation models, such as fully convolutional network (FCN) \cite{long2015fcn}, U-Net \cite{ronneberger2015unet}, and DeepLabs \cite{chen2017deeplab,chen2018deeplabv3plus}, to classify each pixel as building or background \cite{ji2018Bfcn,zhang2020Bfcn,guo2020Bunet,zhu2020Bunet,wang2020Bdeeplab}. However, semantic segmentation-based methods can not distinguish individual buildings. Therefore, some building extraction methods based on instance segmentation models \cite{wen2019Binst,wu2020Binst,xu2021Binst,shi2021Binst} have been studied for building instance segmentation. All these pixel-wise segmentation-based methods fail to obtain accurate building boundaries due to dense buildings and similar backgrounds in remote sensing images. To refine blurred boundaries, some studies \cite{bischke2019BB,zhu2020BB,li2021BB,chen2021BB} introduce boundary-preserved modules to regularize building boundaries. Although recent pixel-wise segmentation methods produce accurate buildings with precise boundaries, they usually output raster building segmentation masks, requiring a delicate post-vectorization pipeline to meet real-world geographic applications.

To vectorize building masks, researchers have formulated building extraction as a multi-stage task and produced vectorized buildings by post-processing or multi-task learning. Early multi-stage methods \cite{zhao2018BPP,wei2019BPP} usually decompose this task into different sub-tasks, including binary building segmentation, polygon generation (or initialization), and boundary regularization. Since these methods are not end-to-end trainable due to separate sub-tasks, building segmentation errors will accumulate throughout the pipeline, resulting in irregular building boundaries. Another line of multi-stage methods \cite{zorzi2021BPP,wei2021BPP,wei2021BPP1,xu2022BPP} has integrated building segmentation, polygonization, and refinement into a framework by multi-task learning. These methods usually design complex pipelines with different threshold constraints for each sub-task, resulting in complex pipelines and hard-to-train.

Recently, dominant building mapping approaches \cite{li2019polymapper,wei2023buildmapper,zorzi2022polyworld} have represented building extraction as polygonal building vertex prediction and directly predicted building vertices to produce vector polygons of buildings. These methods can be categorized as follows: (1) Predict serialized vertices clockwise or counterclockwise from the building feature map of the candidate building region aligned from remote sensing image features. A CNN-RNN architecture is adopted to extract feature maps using convolution neural networks (CNNs) and predict serialized vertices iteratively using recurrent neural networks (RNNs, \textit{e.g.}, ConvLSTM \cite{shi2015convlstm}). Since they output a vector polygon vertex by vertex, this type of method is sensitive to buildings with complex structures or a large number of vertices due to the long dependency problem. (2) Predict a vertex set from the extracted feature map of remote sensing images. The primary concern of these methods is determining the vertices of one building and the vertex sequence of each building from the unordered vertex set, requiring complex human-crafted polygonal constraints and generating irregular building polygons.

In this paper, we present an end-to-end building mapping method with a hierarchical transformer (HiT), which is built on a two-stage detection architecture by adding a polygon head to produce vector polygons of buildings. In particular, the polygon head casts polygonal building as a bidirectional vertex sequence without start or end vertices hypothesis, making serialized vertices prediction order-independent (i.e., the order can be clockwise or equivalently counterclockwise). Through this new perspective, the polygon head pays more attention to the prediction of vertex position and the relationship between two vertices, regardless of whether the order is clockwise or counterclockwise. Unlike RNNs that first define the start vertex and then predict serialized vertices one by one, HiT adopts learnable and order invariant vertex queries to automatically predict serialized vertices of a building at one time. Moreover, we introduce a hierarchical attention mechanism of vertex and edge levels to encode the building feature of the candidate building region aligned from remote sensing image features, providing more geometric information of building boundaries and corners to embed into the building feature.

We evaluate our proposed HiT on two building benchmarks, including the CrowdAI \cite{mohanty2020crowdai} and Inria Polygonized \cite{girard2021ffl} datasets. Since the building mapping task can be seen as an instance segmentation task and a vector polygon extraction task, we compare the proposed HiT with classical instance segmentation methods and state-of-the-art polygonal building extraction methods to evaluate pixel-level and geometric-level performance. Finally, experimental results demonstrate that HiT improves performance to a new state-of-the-art by considerable margins on the two benchmarks.

\begin{figure*}[htbp]
\centering
\includegraphics[width=\linewidth]{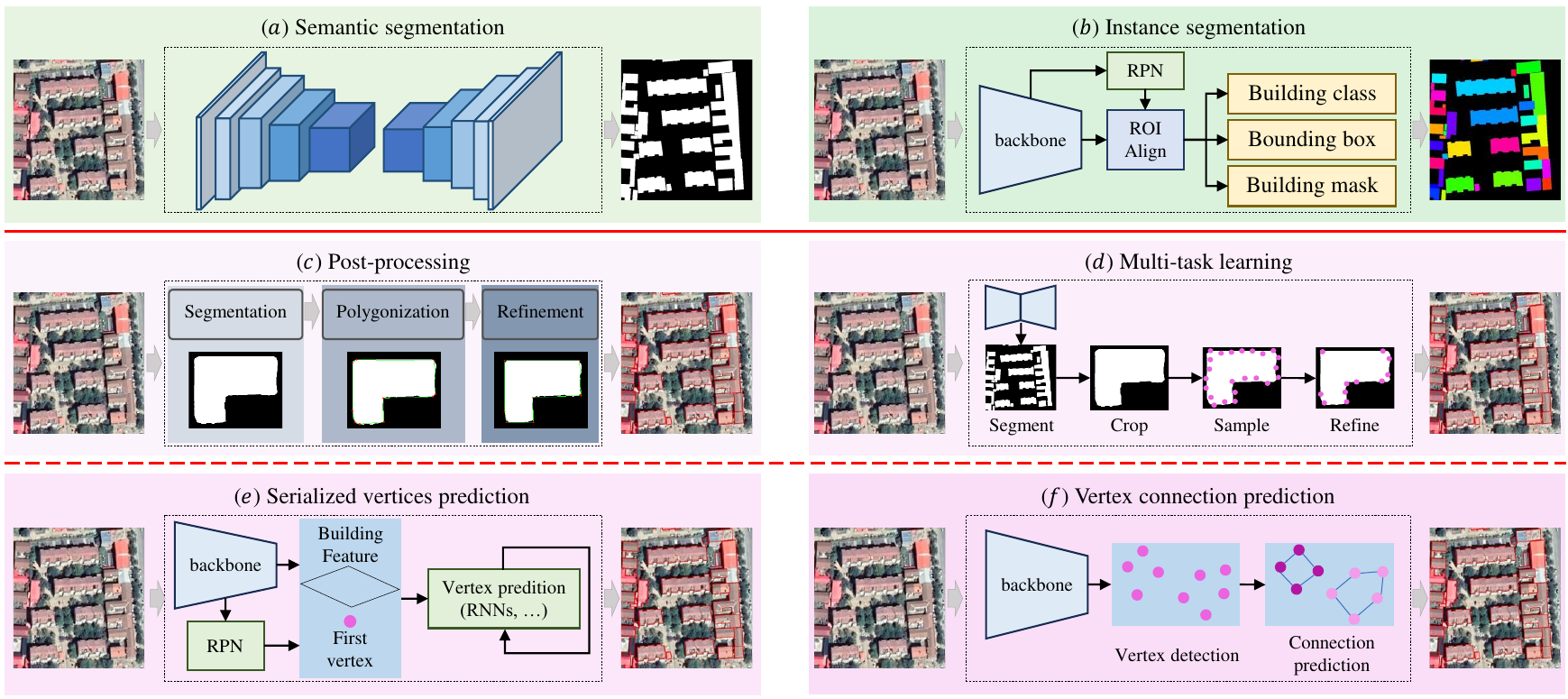}
\caption{Different building mapping categorized into rasterized and polygonal mapping based on the output format. Rasterized mapping employs semantic or instance segmentation frameworks to obtain pixel-wise buildings shown in (a) and (b). Polygonal mapping is subdivided into multi-stage and single-stage pipelines based on whether to segment buildings explicitly. Multi-stage mapping typically adopts post-processing or multi-task learning for transforming pixel-wise masks to polygonal buildings shown in (c) and (d). Single-stage mapping designs serialized vertices or vertex connection prediction modules to obtain building serialized vertices directly shown in (e) and (f).}
\label{fig:metcls}
\end{figure*}

The contributions of our work include: 
\begin{enumerate}
    \item We propose HiT, a two-stage model with three parallel heads to simultaneously detect buildings and extract vector polygons of buildings from remote sensing images, which includes classification, bounding box regression, and polygon heads. HiT is end-to-end trainable and simple yet powerful to achieve building mapping.
    \item HiT represents a polygonal building as a bidirectional vertex sequence, making serialized vertices of a building order-agnostic. In particular, the polygon head applies a transformer-based architecture to directly produce serialized vertices of a building at one time rather than one by one. Besides, we introduce a novel bidirectional polygon loss to supervise the polygon head, avoiding complex polygonal constraints and improving the generalization ability.
    \item We introduce a hierarchical attention mechanism of vertex and edge levels in the encoder of the polygon head, embedding more geometric information (\textit{e.g.}, building boundaries and corners) into aligned building features. Finally, comprehensive experimental results on two building benchmarks demonstrate that our method achieves new state-of-the-art performance on instance segmentation and polygonal building extraction.
\end{enumerate}

The remainder of this paper is organized as follows: Section II reviews related studies. Section III introduces our proposed HiT in detail. Section IV describes experiment settings, including comparison datasets, methods, and evaluation metrics, then reports and discusses the experimental results quantitatively and qualitatively. Section V concludes this work.

\section{Related Work}
Recently, deep learning has been a prevalent technology for remote sensing mapping \cite{su2022nsckl,guo2023dual}. Since building mapping has been a hot research topic in the remote sensing community, many attempts based on deep learning have been widely explored. Early work treats building extraction as a semantic segmentation task \cite{guo2020Bunet,liu2021Bunet,zhu2020Bunet,ji2018Bfcn,zhang2020Bfcn,mou2018Bfcn,wang2020Bdeeplab} or an instance segmentation task \cite{wen2019Binst,wu2020Binst,shi2021Binst,xu2021Binst,li2022Binst,li2023Binst}, as shown in Figure \ref{fig:metcls}(a) and (b). However, they typically output raster building segmentation masks and are not suitable for real-world applications. Recently, polygonal building extraction has directly outputted vector polygons of buildings, which is more suitable for real-world geographic and mapping applications. Therefore, we review literature closely related to our research in this section.

\subsection{Multi-stage polygonal building mapping}
Multi-stage polygonal building mapping first extracts binary building masks and then obtains polygonal buildings using post-processing or multi-task learning, as exemplified in Figure \ref{fig:metcls}(c) and (d). Some multi-stage methods \cite{zhao2018BPP,wei2019BPP,xie2020refined} based on post-processing decomposes polygonal building mapping into sequential sub-tasks: (1) Extract individual building masks by segmentation models (\textit{e.g.}, Mask RCNN \cite{he2017maskrcnn}); (2) Generate or initialize polygons by heuristic post-processing (\textit{e.g.}, Marching Cubes algorithm \cite{we1987marching}); (3) Regularize or simplify polygons using Douglas–Peucker algorithm \cite{douglas1973DPalg} to refine boundaries or vertices. Since they are not end-to-end trainable, building segmentation errors of the first stage will be accumulated throughout the pipeline, resulting in sub-optimal performance and irregular buildings.

To tackle these problems, some works adopt multi-task learning for polygonal building mapping. \cite{wei2021BPP,wei2021BPP1,zorzi2021BPP,xu2022BPP,hu2023BPP} have emerged by integrating building segmentation, polygonization, and refinement into a unified framework to obtain polygonal buildings. FrameField \cite{girard2021ffl} generates a frame field to provide structural information and then aligns the frame field to raster building segmentation for building polygonization. FrameField leverages multi-task learning to achieve a polygonization algorithm utilizing the frame field along with the raster segmentation. \cite{ling2019polygcn,xu2022hisup,li2023joint,khomiakov2023polygonizer} first sample serialized vertices from building masks and then refine vertex positions using the designed refinement module. BuildMapper \cite{wei2023buildmapper} is an end-to-end learnable building contour extraction framework with a learnable contour initialization module and a contour evolution module, which can directly extract building polygons. These models include complex modules with different threshold constraints to achieve each sub-task, making training challenging and computationally intensive. When dealing with inconsistent remote sensing images caused by complex imaging conditions, these methods remain the challenge of performance degradation. Additionally, these methods generally predict a fixed vertex number, resulting in vertex redundancy and insufficiency for different buildings.

\subsection{Single-stage polygonal building mapping}
Single-stage polygonal building mapping casts polygonal building mapping as extracting serialized vertices of a building. As shown in Figure \ref{fig:metcls}(e), \cite{li2019polymapper, zhao2021polymapper++}, motivated by PolyRNNs \cite{castrejon2017polyrnn,acuna2018polyrnn++}, first extracts building features and then iteratively predicts building serialized vertices by RNNs. TransBuilding \cite{zhang2023transbuilding} predicts polygonal buildings with a vertex transformer module and designs three self-attention modules in row-wise, column-wise, and vertex-wise to enhance geometric information of building features. These methods typically employ a two-stage detection framework (\textit{i.e.}, Faster RCNN \cite{girshick2015fasterrcnn}) and add a serialized vertices prediction head parallel with building classification and bounding box regression heads. Figure \ref{fig:metcls}(f) shows another line of works, which first detects all the vertices and then predicts the vertex connection matrix for assembling serialized vertices. PolyWorld \cite{zorzi2022polyworld} directly predicts a connection matrix to find the vertices of one building and the order of building vertices. Since PolyWorld \cite{zorzi2022polyworld} produces serialized vertices in a bottom-up pathway, missing or error vertices will influence the connection matrix learning, leading to self-intersection or non-closed polygons.

\subsection{Transformers in CV}
Since Transformers \cite{vaswani2017transformer} are successful in natural language processing (NLP) with their powerful feature encoding ability, some researchers have extended them to computer vision (CV). Self-attention mechanism enables Transformers to model long dependencies, and multiple attention heads learn appropriate inductive bias, avoiding spatial constraints and inductive bias in convolutional operations. Hence, Transformers have achieved promising performance over CNN-based approaches \cite{han2022survey} in CV. ViT \cite{dosovitskiy2020ViT} splits an image into non-overlapping patches and employs the standard transformer-based structure to process sequences of image patches, which has become a milestone work in vision transformers. Since ViT is a plain and non-hierarchical network, Swin Transformer \cite{liu2021swin} introduces a feature pyramid to extract multi-scale feature maps, which serves as a hierarchical backbone and facilitates Transformers in other tasks. Follow-up works \cite{wu2021cvt,wang2022pvt,liu2022swin,xia2022vision} adopt hierarchical stages with spatial reduction layers and hybrid architectures with convolutional operations to efficiently extract local and global information. With notably advanced vision transformers have emerged in image classification, transformers have been successfully applied in various fields, such as detection \cite{carion2020Detr}, segmentation\cite{zheng2021rethinking}, and video \cite{wang2021end} in CV. DETR \cite{carion2020Detr} is the first transformer-based detection framework, representing object detection as a set prediction and matching problem and removing additional operations such as anchor generation and non-maximum suppression (NMS). SETR \cite{zheng2021rethinking} reformulates semantic segmentation as a sequence-to-sequence prediction task and uses a pure Transformer to model the global context in transformer layers, which can provide a powerful segmentation model. This line of work formulates detection or segmentation problems as set prediction tasks and introduces learnable queries to extract targets in an auto-regressive manner, which can be applied in vertex prediction. Alfieri et al. \cite{alfieri2021polytrans} explore transformer-based architecture in polygon prediction, but it still needs a multi-layer Elman RNN \cite{sherstinsky2020ElmanRNN} to generate serialized vertices iteratively. However, no prior work has exploited Transformer for serialized vertex prediction. This work leverages a pure Transformer to predict serialized vertices for polygonal building mapping, aiming to mitigate the research gap.

\begin{figure*}[htbp]
\centering
\includegraphics[width=0.85\linewidth]{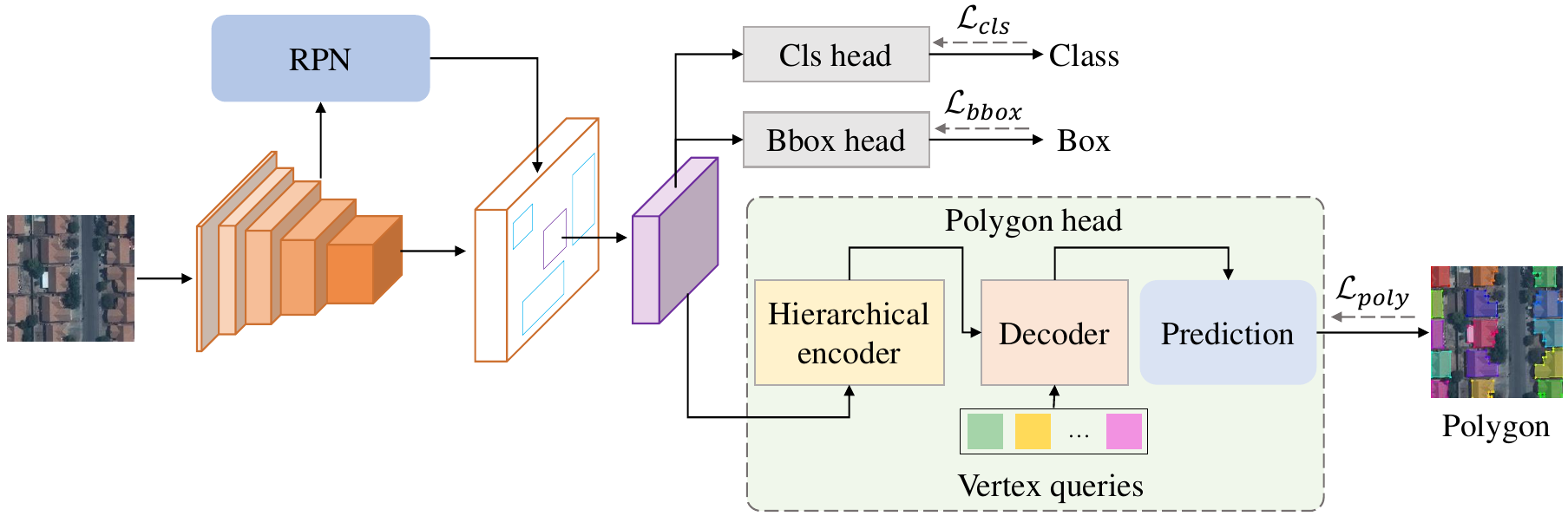}
\caption{Overview of HiT. HiT is a two-stage building mapping framework, which includes classification, bounding box regression, and polygon heads. The polygon head predicts serialized vertices of a building, together with building detection. We introduce a novel bidirectional polygon loss to train the polygon head without complex constraints.}
\label{fig:HiT}
\end{figure*}

\section{Method}
In this section, We introduce the proposed HiT, a single-stage polygonal building mapping approach. In the following, we will first describe the overall pipeline of the proposed HiT and then describe each component in detail.

\subsection{Overall pipeline}
Polygonal building mapping has recently focused on iteratively predicted vertex on the condition of the predicted first vertex and the previous predicted vertices. Due to the long dependency problem, it remains very challenging to handle buildings with complex structures or occlusions and shadows caused by the imaging conditions. In response to these challenges, HiT models the vertex sequence on the condition of all the vertices by the designed polygon head. Specifically, we build the polygon head based on the insight that building polygons should be effectively delineated using a bidirectional vertex sequence. This innovative perspective makes serialized vertex prediction order-independent, enabling the model to concentrate on predicting vertex positions and relationships of any two adjacent vertices. Consequently, this alleviates the reliance on assumptions about the starting or ending vertices within the single-stage pipeline.

As shown in Figure \ref{fig:HiT}, HiT employs a transformer-based architecture to simultaneously predict serialized vertices. Notably, the polygon head incorporates learnable and order-invariant vertex queries to dynamically predict serialized vertices for buildings with varying numbers at one time. Furthermore, HiT introduces a hierarchical attention mechanism at the vertex and edge levels, integrating convolution operations to amplify the encoding of geometric information for building features. A novel bidirectional polygon loss is further introduced to supervise the polygon head, thereby improving the learning of sequence relationships within the query-based transformer module. In addition, the designed loss disregards the clockwise or counterclockwise orientation of the sequence, enhancing greater flexibility when predicting polygon vertex sequences. Finally, HiT directly extracts building polygons with appropriate vertices, achieving high performance compared to multi-stage and single-stage polygonal building mapping methods detailed in subsection \ref{subsec:resdis}.

\begin{figure*}[htbp]
\centering
\includegraphics[width=0.8\linewidth]{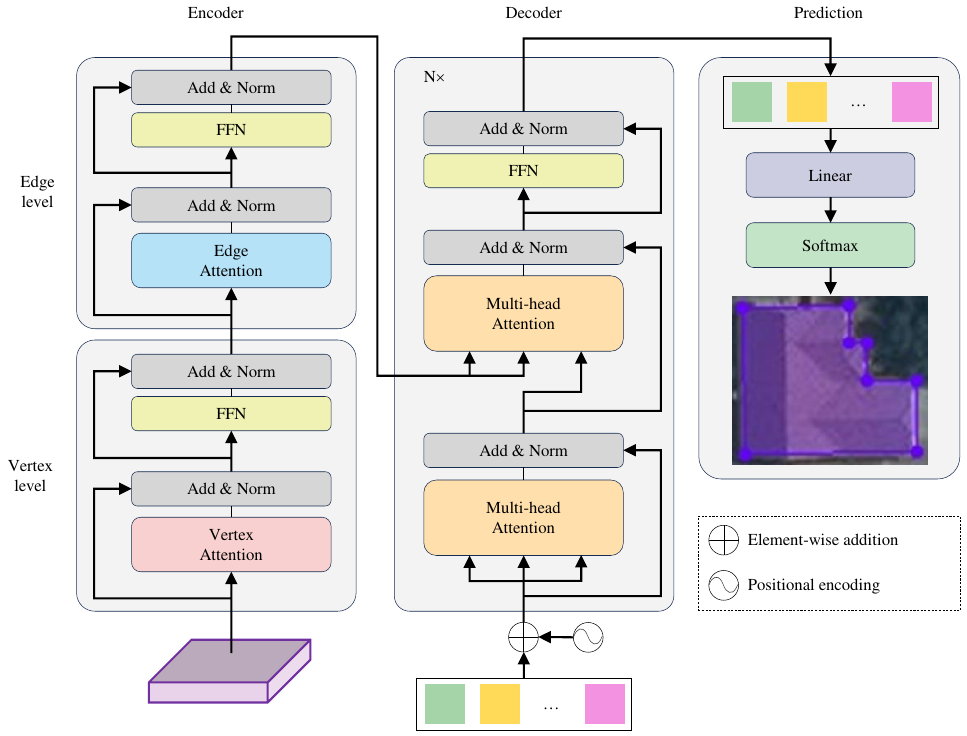}
\caption{Illustration of the polygon head. The encoder with a hierarchical attention mechanism embeds more geometric information into the building feature. The decoder learns vertex queries to predict serialized vertices.}
\label{fig:polygon}
\end{figure*}

\subsection{Building detection}
In building detection, HiT first extracts multi-scale features from the input image by the feature extraction module and then detects buildings by classification and regression heads. The feature extraction module consists of multi-scale feature extraction and candidate building region generation. Following Faster RCNN \cite{girshick2015fasterrcnn}, HiT takes the ResNet50 \cite{he2016resnet} as the multi-scale backbone network in the feature extraction module. In this paper, multi-scale features $\textit{C}_i$ ($i\in[1, 2, 3, 4, 5]$) are firstly extracted from a remote sensing image $\textit{X} \in \mathbb R^{3\times H_0\times W_0}$ by ResNet-50, of which the feature channels are \{64, 256, 512, 1024, 2048\} and the resolutions are \{1/2, 1/4, 1/8, 1/16, 1/32\} of the input image $\textit{X}$. In order to accurately detect buildings of different sizes, an FPN fuses multi-scale features \{$\textit{C}_2, \textit{C}_3, \textit{C}_4, \textit{C}_5$\} in a top-down pathway with the lateral connection, which obtains new multi-scale features $\textit{P}_i$ ($i\in[2, 3, 4, 5, 6]$) with 256 channels and \{1/4, 1/8, 1/16, 1/32, 1/64\} resolutions of the input image $\textit{X}$. Finally, an RPN is adopted to generate candidate building regions with three anchor aspect ratios \{0.5, 1.0, 2.0\} from fusion features \{$\textit{P}_2, \textit{P}_3, \textit{P}_4, \textit{P}_5, \textit{P}_6$\}.

The building detection module outputs building classification scores and bounding boxes through the building classification head and the bounding box regression head. Given candidate building regions from the RPN, the building detection module first extracts the building feature map $\textit{B} \in \mathbb R^{256\times 7\times 7}$ from multi-scale features $\textit{P}_i$ of the corresponding scale through a ROIAlign operation \cite{he2017maskrcnn}. Subsequently, the building feature map \textit{B} is flattened along the spatial and channel dimensions and goes through two connected linear layers to reduce channel dimension, which obtains the instance-level feature representation. Finally, a classification linear layer takes as input the building representation and predicts the building score (\textit{i.e.}, building or background); on the other hand, a bounding box regression linear layer inputs the building representation and produces the building bounding box including center, width, and length.

\subsection{Polygon prediction}
HiT represents the vector polygon of a building as a bidirectional vertex sequence, of which the vertex order can be clockwise or equivalently counterclockwise. Therefore, the polygonal building mapping can be formulated as a sequence prediction task. As shown in Figure \ref{fig:polygon}, HiT introduces a transformer-based polygon head to directly predict serialized vertices of a building at one time. Firstly, the polygon head also uses a ROIAlign operation \cite{he2017maskrcnn} to extract the corresponding building feature map $\textit{B} \in \mathbb R^{256\times 20\times 20}$, which is a large resolution compared to the building detection. Then, the encoder of the polygon head adopts a hierarchical attention mechanism to embed geometric information into the building feature map $\textit{B}$. Finally, the decoder of the polygon head learns dynamic vertex queries from the building embedding for automatically serialized vertices prediction.

\noindent \textbf {Encoder}. 
The encoder of the standard transformer architecture exploits the self-attention operation to calculate the similarity between every two tokens of the sequence. However, serialized vertices are sparsely present in the building feature map $\textit{B}$, leading to the self-attention operation intensive computation, memory cost, and low efficiency. To deal with the sparsity of attention weights, the encoder of the polygon head replaces the original self-attention mechanism with the hierarchical attention mechanism to encode the building feature map $\textit{B}$ efficiently, which avoids the complexity and speeds up the convergence speed by introducing the geometric information in terms of vertex and edge levels.

\begin{figure}[!tp]
\centering
\includegraphics[width=0.6\linewidth]{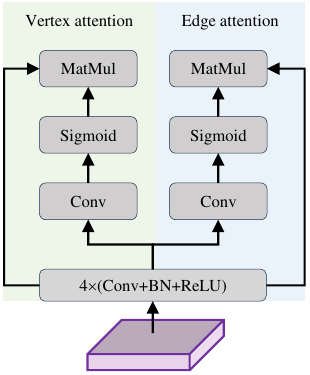}
\caption{Illustration of vertex-level and edge-level attention operations. Vertex-level and edge-level attention replace the original self-attention mechanism to encode the building feature map, avoiding the complexity and speeding up the convergence speed by introducing the geometric information in terms of vertex and edge levels.}
\label{fig:v_e_attn}
\end{figure}

As depicted in Figure \ref{fig:polygon}, the encoder consists of vertex-level and edge-level blocks. Similar to the original encoder in the transformer \cite{vaswani2017transformer}, the two blocks have an attention operation and a feed-forward network ($\text{FFN}$), after which the short-cut connection and the layer normalization are also added. Since HiT adopts the hierarchical attention mechanism, the encoder directly takes the building instance feature $\textit{B}$ as input and outputs the building embedding, avoiding feature patchy and positional encoding. As shown in Figure \ref{fig:v_e_attn}, the encoder first uses four $3\times3$ convolution layers with the batch normalization and a rectified linear unit (ReLU \cite{glorot2011relu}) to process the building feature map $\textit{B}$. Then, a $1\times1$ convolution layer and a sigmoid activation function are used to obtain the vertex attention weight, and another $1\times1$ convolution layer, followed by a sigmoid operation, is used to obtain the edge attention weight. Finally, the vertex-level attention is calculated by multiplying the building feature map $\textit{B}$ and the vertex attention weight. Similarly, the edge-level attention is calculated by multiplying the building feature map $\textit{B}$ and the edge attention weight. The outputs of the vertex-level and edge-level attention are defined as:

\begin{equation} \label{eq:evattn}
  \begin{aligned}
  \textit{B}^{'} &= 4*[\text{ReLU}(\text{BN}(\text{conv}(B)))]              \\
  \textit{Attn}_v &= \textit{B} \otimes \sigma(\text{conv}(\textit{B}^{'})) \\
  \textit{Attn}_e &= \textit{B} \otimes \sigma(\text{conv}(\textit{B}^{'}))
  \end{aligned}
\end{equation}

Afterward, the FFN is added to generate the final building embedding $\textit{B}_{emd}$, which can increase the expressive ability of the model. In this paper, we integrate the information from vertex to edge level, enhancing building features. Besides, we will discuss the combination manner of vertex-level and edge-level attention in ablation studies.

\noindent \textbf {Decoder}. 
Unlike RNNs, which iteratively predict serialized vertices of a building, the decoder uses the vertex query $\textit{Q} \in \mathbb R^{M\times256}$ to predict the vertex sequence at one time. Like the original decoder of the standard transformer architecture, the decoder consists of $N$ identical decoder blocks to perform multi-head self-attention and cross-attention. Each block includes a multi-head self-attention sub-layer, a cross-attention sub-layer, and an FFN sub-layer. Besides, the short-cut connection and the layer normalization are added after attention operations and $\text{FFN}$.

As shown in Figure \ref{fig:polygon}, the vertex query $\textit{Q}$ added with the sinusoidal positional encoding is transformed by the multi-head self-attention operation, which outputs the vertex embedding followed by the short-cut connection and the layer normalization operations. Subsequently, the cross-attention sub-layer ($\text{CA}_{v}$) inputs the building embedding from the encoder as key and value and uses the output vertex embedding to automatically integrate the information of serialized vertices, after which the short-cut connection and the layer normalization operations are also used to enhance the output vertex embedding. Finally, the FFN sub-layer is applied with the short-cut connection and the layer normalization operations to generate the final vertex embedding. The output of each block is defined as:

\begin{equation} \label{eq:decoder}
  \begin{aligned}
  \textit{V}^{'} &= \textit{Q} \oplus \text{PE}(\textit{Q}) \\
  \textit{V}_{emd} &= \text{LN}(\text{SA}_{v}(\textit{V}^{'}) \oplus \textit{V}^{'}) \\
  \textit{V}_{emd} &= \text{LN}(\text{CA}_{v}(\textit{B}_{emd}, \textit{B}_{emd}, \textit{V}_{emd}) \oplus \textit{V}_{emd}) \\
  \textit{V}_{emd} &= \text{LN}(\text{FFN}(\textit{V}_{emd}) \oplus \textit{V}_{emd})
  \end{aligned}
\end{equation}
where $\text{PE}$ is the positional encoding. $\text{SA}_{v}$ and $\text{CA}_{v}$ are the multi-head self-attention and the multi-head cross-attention operations. $\text{LN}$ is the layer normalization operation. $\oplus$ means the element-wise addition. In the prediction, the final vertex embedding $\textit{V}_{emd}$ is fed to a linear layer and is transformed to \textit{M} one-hot vectors by the softmax operation, indicating whether the vertex is a building vertex or not.

\subsection{Training objective}
\noindent \textbf {Building detection}. 
HiT consists of building classification and bounding box regression. Building classification loss $\textit{L}_{cls}$ is calculated by the binary cross-entropy loss:
\begin{equation} \label{eq:cls}
  \begin{aligned}
  \textit{L}_{cls} &= -\frac{1}{N}\sum_{i = 1}^{N}(y_i \cdot \text{log}(p_i)+(1-y_i) \cdot \text{log}(1-p_i))
  \end{aligned}
\end{equation}
where $y_i$ represents the class, which is 1 for the building class or 0 for the background, and $p_i$ is the predicted classification score. For building bounding box regression loss $\textit{L}_{bbox}$, building detection is trained using a L1 loss:
\begin{equation} \label{eq:bbox}
  \begin{aligned}
  \textit{L}_{bbox} &= \frac{1}{N}\sum_{i = 1}^{N}\vert t_i - t_i^* \vert
  \end{aligned}
\end{equation}
where $t_i$=$(cx_i,cy_i,w_i,h_i)$ and $t_i^*$=$(cx_i^*,cy_i^*,w_i^*,h_i^*)$ represent the ground truth and predicted bounding boxes, respectively.

\noindent \textbf {Polygon prediction}. 
Since polygon prediction introduces geometric constraints in terms of vertex-level and edge-level, the vertex and edge prediction is trained by the focal loss \cite{lin2017focalloss}:
\begin{equation} \label{eq:ver_edge}
  \begin{split}
  \textit{L}_{ver}&=
        \begin{cases}
            -\alpha(1-p^v)^{\gamma}\text{log}(p^v)& \text{vertex}\\
            -\alpha(p^v)^{\gamma}\text{log}(1-p^v)& \text{otherwise}
        \end{cases} \\
  \textit{L}_{edge}&=
        \begin{cases}
            -\alpha(1-p^e)^{\gamma}\text{log}(p^e)& \text{edge}\\
            -\alpha(p^e)^{\gamma}\text{log}(1-p^e)& \text{otherwise}
        \end{cases}
  \end{split}
\end{equation}
where $\alpha$ and $\gamma$ are the hyper-parameters, which is 2.0 and 4.0 in this paper. $p^v$ and $p^e$ denotes the predicted probability of vertex and edge, respectively.

\begin{figure}[!tp]
\centering
\includegraphics[width=\linewidth,scale=1.00]{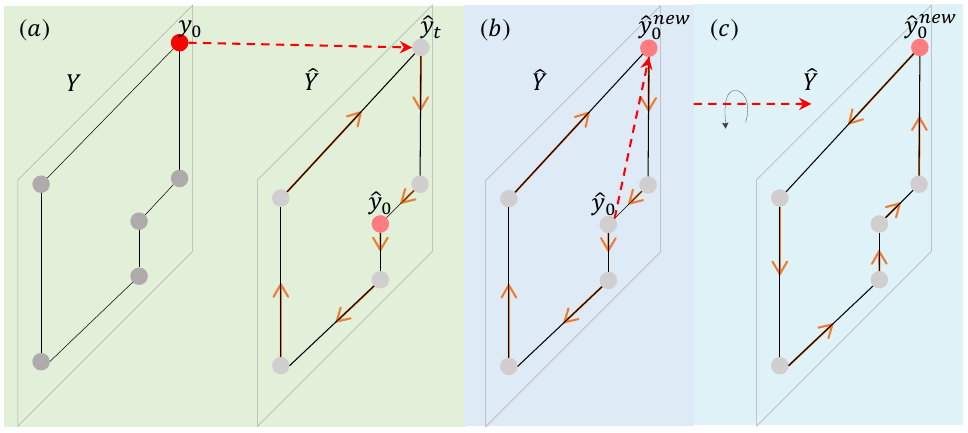}
\caption{Illustration of the serialized vertices prediction loss $\textit{L}_{sv}$. (a)Search the corresponding vertex. (b)Shift the predicted serialized vertices. (c)Inverse the predicted serialized vertices.}
\label{fig:svloss}
\end{figure}

For serialized vertices prediction loss $\textit{L}_{sv}$, we use the binary cross-entropy loss to automatically predict a polygon with orientation invariant. In specific, we define $\bf{Y}$ and $\hat{\bf{Y}}$ as the ground truth and predicted serialized vertices, respectively. For ease of understanding, we describe the calculation of $\textit{L}_{sv}$ by using the first vertex $y_0$ of $\bf{Y}$. As illustrated in Figure \ref{fig:svloss}, we first search the reference vertex ${\hat{y}_t}$ from $\hat{\bf{Y}}$, which is closest to $y_0$. Subsequently, we move ${\hat{y}_t}$ to the first vertex ${\hat{y}_0^{new}}$ by sequentially shift the predicted sequence. Then, we fix ${\hat{y}_0^{new}}$ and inverse the other vertices in counter clockwise to get ${\hat{\bf{Y}}}_{inv}$. Finally, we calculate the minimum of the binary cross-entropy loss between ground truth $\bf{Y}$ and $\hat{\bf{Y}}$, as well as ${\hat{\bf{Y}}}_{inv}$. The $\textit{L}_{sv}$ is defined as follows:
\begin{equation} \label{eq:sv}
  \begin{split}
  \hat{y}_t&=\text{Search}(\hat{\bf{Y}}, y_0) \\
  \hat{\bf{Y}}&=\text{Shift}(\hat{\bf{Y}}, \hat{y}_t) \\
  \hat{\bf{Y}}_{inv}&=\text{Inverse}(\hat{\bf{Y}}) \\
  \textit{L}_{sv}&=\text{min}(\text{L}_{ce}(\bf{Y}, \hat{\bf{Y}}), \text{L}_{ce}(\bf{Y}, {\hat{\bf{Y}}}_{inv}))
  \end{split}
\end{equation}
The polygon prediction loss $\textit{L}_{poly}$ is the the sum of $\textit{L}_{ver}$, $\textit{L}_{edge}$, and $\textit{L}_{poly}$. 
Finally, the total loss of HiT is defined as:
\begin{equation} \label{eq:sv}
  \begin{split}
  \textit{L}&=\textit{L}_{cls}+\textit{L}_{bbox}+\textit{L}_{poly}
  \end{split}
\end{equation}

\subsection{Implementation Details}
HiT is implemented based on Faster RCNN \cite{girshick2015fasterrcnn} with a ResNet50 backbone, which is trained end-to-endly by using the PyTorch framework. For data augmentations during training and inference, input images are generally resized to 512$\times$512, randomly flipped with a probability of 0.5, and normalized. During training, HiT is optimized by AdamW \cite{loshchilov2018adamw} optimizer. The initial learning rates of the backbone and the other are set to 1e-5 and 1e-4, respectively. The weight decay is set to 1e-4. The model is trained for 150 epochs with the learning rate dropped by 10 at the 90 and 130 epochs. For the model hyper-parameters and the joint training strategy, we have conducted extensive ablation studies, as illustrated in the section \ref{sec:exp_ablation}.

\section{Experiments}

\subsection{Dataset}
The proposed HiT is evaluated on two public building segmentation benchmark datasets, namely the CrowdAI Mapping Challenge dataset (CrowdAI dataset) \cite{mohanty2020crowdai} and the Inria Aerial Image Labeling dataset (Inria dataset) \cite{maggiori2017inria}, to assess its performance and generalization. Buildings in the two large-scale building datasets cover many regions with different complex scenes and significantly vary in size, shape, structure, and appearance.

\textit{(1) CrowdAI dataset}: 
The CrowdAI dataset is a large-scale satellite imagery of about 30 cm resolution with RGB channels, in which images have a size of 300 $\times$ 300 pixels and are annotated with polygonal building instances in MS-COCO \cite{lin2014mscoco} format. The training set consists of 280,741 images with around 2,400,000 polygonal building instances. The test set has 60,317 images with 515,364 polygonal building instances. In addition, a small version that only includes 8,366 images for the training set and 1,820 images for the test set is also provided for comparison experiments. In our paper, the small version is used to conduct the following ablation studies to consider time-consuming and resource constraints.

\textit{(2) Inria dataset}: 
The Inria dataset contains 180 aerial images of 5,000 $\times$ 5,000 pixels, covering different geographic locations (\textit{i.e.}, United States and Austria) ranging from highly dense metropolitan financial districts to alpine resorts. Buildings of the Inria dataset have different urban settlement appearances and are annotated in binary masks, indicating pixels into building and not building classes. The Inria dataset has a spatial resolution of 30 cm and is split by the cities for the training set and the test set for assessing the model's generalization. 

Since the source annotations in the Inria dataset are pixel-wise semantic masks, the Inria dataset is not suitable for supervising the model to extract polygonal building instances. Following the FrameField \cite{girard2021ffl}, we use the Inria Polygonized dataset to train our model, which converts the source annotations to polygonal MS-COCO \cite{lin2014mscoco} format. The images of the Inria Polygonized dataset are cropped into 512 $\times$ 512 patches with an overlap of 128. In the cropping stage, we remove buildings with an area smaller than 50\% compared to the original building instance. Finally, we split the cropped images with the 75\% for the training set and the 25\% for the test set.

\subsection{Evaluation Metrics}
Building mapping from remote sensing images is a building instance segmentation task and a polygonal building extraction task. Therefore, we adopt two evaluation criteria to compare the proposed method with other methods.

\noindent \textbf {Instance metric}. 
For the instance segmentation, we adopt the average precision (AP) and the average recall (AR) under different intersection over union (IoU) thresholds provided by the standard MS-COCO metrics \cite{lin2014mscoco}. In order to evaluate the overall performance, we use $AP$ and $AR$ metrics, which present average precision and average recall under IoU thresholds ranging from 0.50 to 0.95 with a step of 0.05. Moreover, $AP_{50}$, $AP_{75}$, $AR_{50}$ and $AR_{75}$ are also calculated under IoU thresholds of 0.5 and 0.75 to measure the model's basic and higher performance. Besides, we report $F1$ to comprehensively assess the model's precision and recall, which is calculated as shown in Eq. \ref{eq:inst}.
\begin{equation} \label{eq:inst}
  \begin{split}
  \textit{AP}&=\frac{AP_{0.50}+AP_{0.55}+ \cdots +AP_{0.95}}{10} \\
  \textit{AR}&=\frac{AR_{0.50}+AR_{0.55}+ \cdots +AR_{0.95}}{10} \\
  \textit{F1}&=\frac{2 \times AP \times AR}{AP+AR} 
  \end{split}
\end{equation}
where $AP_i$ and $AR_i$ measure average precision and average recall under IoU threshold $i$ that is calculated by $IoU =  (Pre \cap GT)/(Pre \cup GT)$.

\noindent \textbf {Polygonal metric}. 
The polygonal metric considers the geometric properties of the extracted buildings, so we adopt three indicators to evaluate the extracted polygonal buildings, which consist of the N ratio \cite{zorzi2022polyworld}, the complexity aware IoU (C-IoU) \cite{zorzi2022polyworld}, and the Max Tangent Angle Error (MTA) \cite{girard2021ffl}. The N ratio measures the simplicity of polygonal buildings by calculating the ratio between the predicted vertex number and the ground truth, which is defined as:
\begin{equation} \label{eq:nratio}
  \begin{split}
  \textit{N ratio}&=\frac{\hat{V}_N}{V_N}
  \end{split}
\end{equation}
where $V_N$ and $\hat{V}_N$ denote the vertex number of the prediction and the ground truth. When the model predicts redundant or insufficient vertices, the N ratio would be greater or less than 1. So, the N ratio is closer to 1, illustrating the better performance of the model. C-IoU is used to jointly assess the complexity and the segmentation of the extracted polygonal buildings, which is defined as:
\begin{equation} \label{eq:nratio}
  \begin{split}
  \textit{RD}&=\frac{\vert V_N-\hat{V}_N \vert}{V_N+\hat{V}_N} \\
  \textit{C-IoU}&=\text{IoU}(M, \hat{M}) \cdot (1-\textit{RD})
  \end{split}
\end{equation}
where M and $\hat{M}$ denote masks of the predicted polygonal building and the ground truth. The C-IoU is higher when the model extracts polygonal buildings with accurate segmentation and precise polygonal complexity. MTA measures geometric shape by calculating the tangent angles between the predicted polygonal building and the ground truth, which is defined as:
\begin{equation} \label{eq:nratio}
  \begin{split}
  \text{T}(V_i) &= (V_{i+1}-V_i)/\Vert V_{i+1}-V_i \Vert \\ 
  \text{T}(\hat{V}_i) &= (\hat{V}_{i+1}-\hat{V}_i)/\Vert \hat{V}_{i+1}-\hat{V}_i \Vert \\
  \textit{MAT} &= \max \limits_{1 \leq x \leq N}\text{cos}^{-1}(\langle \text{T}(V_i), \text{T}(\hat{V}_i) \rangle)
  \end{split}
\end{equation}

\begin{table*} [htbp]
\centering
\caption{Results on the CrowdAI dataset under the instance segmentation metric. The best results are marked in bold.\label{tab:crowdai_inst}}
\begin{tabular}{lccccccc}
\hline
Method     & AP $\uparrow$ & $AP_{50}$ $\uparrow$ & $AP_{75}$ $\uparrow$ & AR $\uparrow$ & $AR_{50}$ $\uparrow$ & $AR_{75}$ $\uparrow$ & $F1$ $\uparrow$ \\
\hline
Mask RCNN \cite{he2017maskrcnn}         & 41.9 & 67.5  & 48.8 & 47.6 & 70.8 & 55.5 & 44.6 \\
PANet \cite{liu2018panet}               & 50.7 & 73.9  & 62.6 & 54.4 & 74.5 & 65.2 & 52.5 \\
PolyMapper \cite{li2019polymapper}      & 55.7 & 86.0  & 65.1 & 62.1 & 88.6 & 71.4 & 58.7 \\
FrameField \cite{girard2021ffl}         & 61.3 & 87.5  & 70.6 & 65.0 & 89.4 & 73.9 & 63.1 \\
PolyWorld \cite{zorzi2022polyworld}     & 63.3 & 88.6  & 70.5 & 75.4 & 93.5 & 83.1 & 68.8 \\
TransBuilding \cite{zhang2023transbuilding} & 54.4 & 88.6  & 64.1 & 62.1 & 91.6 & 72.7 & 56.0 \\
BuildMapper \cite{wei2023buildmapper}   & 63.9 & 90.1  & 75.0 & - & - & - & - \\
HiT (ours) & $\textbf{64.6}$ & $\textbf{91.9}$ & $\textbf{78.7}$ & $\textbf{75.5}$ & $\textbf{93.8}$ & $\textbf{83.5}$ & \textbf{69.6} \\
\hline
\end{tabular}
\end{table*}

\begin{table} [!tb]
\centering
\caption{Results on the CrowdAI dataset under the polygonal metric. ``N ratio (=1)'' denotes that the performance is better when the N ratio is closer to 1. The best results are marked in bold.\label{tab:crowdai_poly}}
\begin{tabular}{lccc}
\hline
Method   & C-IoU $\uparrow$ & MTA $\downarrow$ & N ratio (=1)  \\
\hline
PolyMapper \cite{li2019polymapper}      & 65.3  & 32.8 & 1.29  \\
FrameField \cite{girard2021ffl}         & 73.7  & 33.5 & 1.13  \\
PolyWorld \cite{zorzi2022polyworld}     & 88.2  & 32.9 & 0.93  \\
HiT (ours)         & $\textbf{88.6}$ & $\textbf{31.7}$ & $\textbf{1.00}$ \\
\hline
\end{tabular}
\end{table}

\subsection{Results and Discussion}
\label{subsec:resdis}
In our experiments, we select PolyMapper \cite{li2019polymapper}, Framefield \cite{girard2021ffl}, and PolyWorld \cite{zorzi2022polyworld}) for comparison, which are recently proposed state-of-the-art (SOTA) methods for polygonal building extraction. Besides, we compare HiT with more polygonal building mapping methods (TransBuilding \cite{zhang2023transbuilding} and BuildMapper \cite{wei2023buildmapper}) on CrowdAI dataset. Moreover, we compare our method with classical instance segmentation methods, including Mask RCNN \cite{he2017maskrcnn} and PANet \cite{liu2018panet} following the recent SOTA methods. We adopt the Douglas-Peucker algorithm \cite{douglas1973DPalg} to polygonize pixel-wise segmentation masks to obtain polygonal results from binary building instance masks predicted by instance segmentation methods (Mask RCNN and PANet), which can be compared with the other polygonal building extraction for fair comparison. Our proposed HiT is similar to Mask RCNN, which all add branches for task-specific prediction in a standard two-stage detection framework (a mask prediction head in the Mask RCNN and a polygon prediction head in the proposed HiT). Therefore, we select the Mask RCNN as the baseline.

\noindent \textit{(1) Results on CrowdAI dataset}.

For the CrowdAI dataset, the proposed HiT is compared with Mask RCNN \cite{he2017maskrcnn}, PANet \cite{liu2018panet}, PolyMapper \cite{li2019polymapper}, FrameField \cite{girard2021ffl}, PolyWorld \cite{zorzi2022polyworld}, TransBuilding \cite{zhang2023transbuilding} and BuildMapper \cite{wei2023buildmapper} under the instance segmentation metric and the polygonal metric, respectively. Since TransBuilding \cite{zhang2023transbuilding} and BuildMapper \cite{wei2023buildmapper} only evaluates performance on the instance segmentation metric, we have excluded them from Table \ref{tab:crowdai_poly}.

\noindent \textbf{Quantitative Evaluation}. 
Table \ref{tab:crowdai_inst} and \ref{tab:crowdai_poly} report the quantitative comparison results under the two metrics. From Table \ref{tab:crowdai_inst}, we can see that the proposed method outperforms all the comparison methods under the instance segmentation metric. Compared with the baseline instance segmentation method, the AP, AR, and F1 scores have been improved by +22.5\%, +27.9\%, and +24.9\%, respectively. These significant improvements show that the designed polygon head can more efficiently extract building instances than the mask head in Mask RCNN. For polygonal segmentation methods, the proposed HiT has achieved AP gains of +8.7\%, +2.7\%, and +1.1\% compared with polygonal building extraction methods (\textit{i.e.}, PolyMapper, FrameField, and PolyWorld). In addition, the AR scores are +13.4\%, +10.1\%, and +0.1\% higher than the three polygonal building extraction methods, respectively. Specifically, the F1 score of the proposed HiT has outperformed the three polygonal building methods by +10.8\%, +6.0\%, and +0.7\%. All these results consistently show that the proposed HiT generates building instances with high precision and recall.  
\begin{figure*}[htbp]
\centering
\includegraphics[width=0.9\linewidth,scale=1.00]{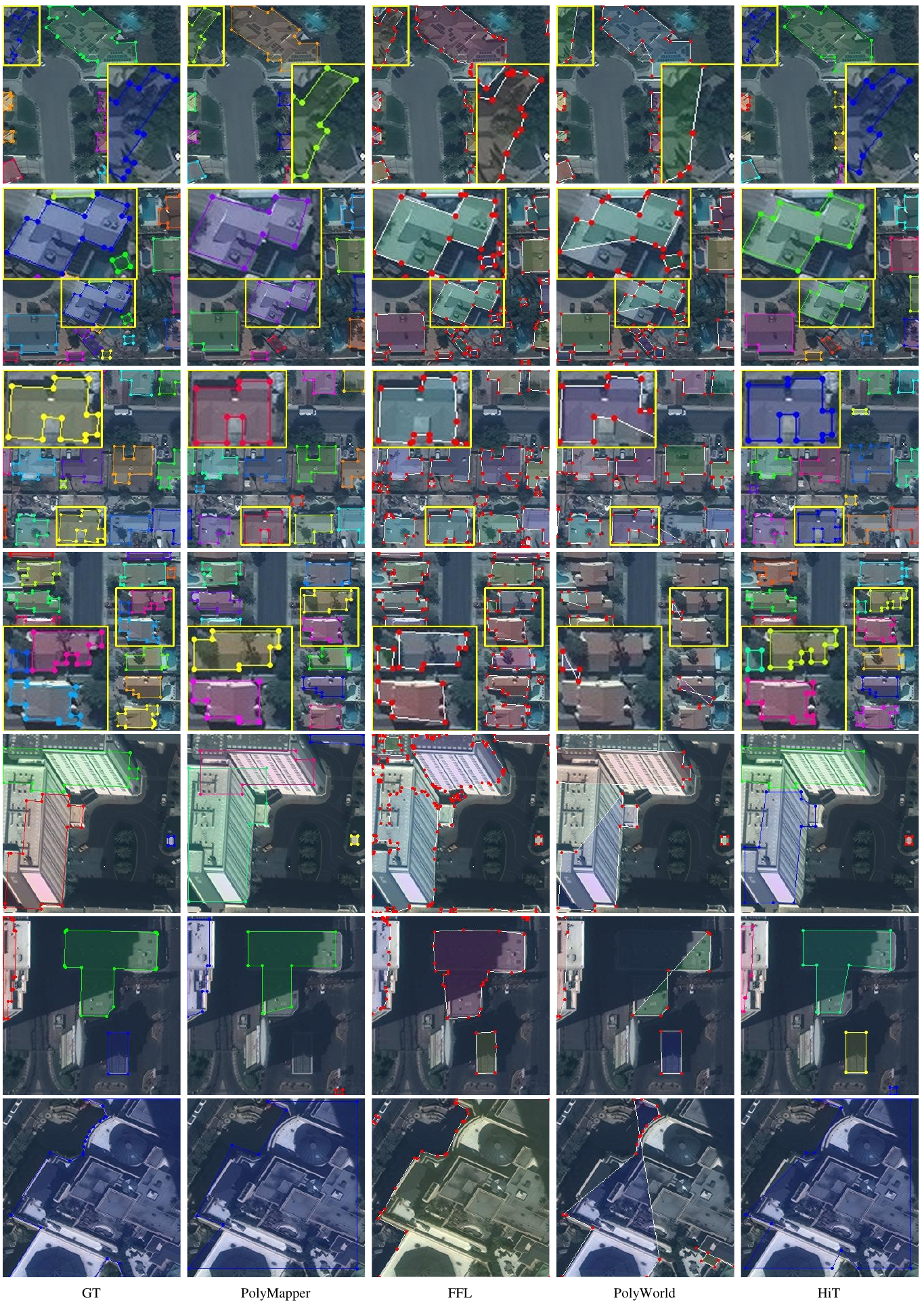}
\caption{Qualitative results on CrowdAI. The proposed HiT can generate high-quality polygonal buildings of different sizes and shapes.}
\label{fig:crowdaires}
\end{figure*}

\begin{figure*}[htbp]
\centering
\includegraphics[width=0.85\linewidth]{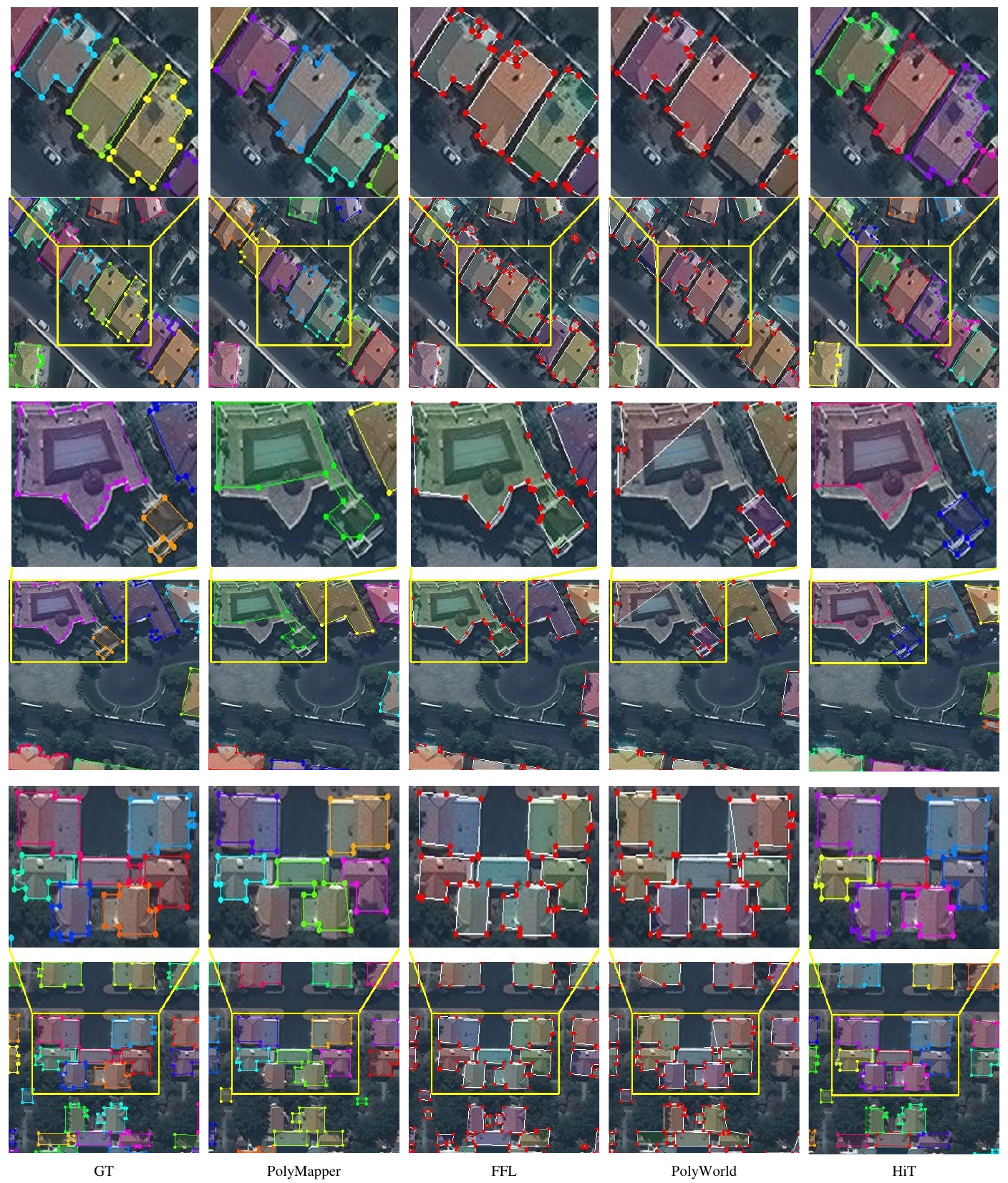}
\caption{Additional qualitative results on CrowdAI. HiT can accurately extract polygonal buildings.}
\label{fig:crowdaires2}
\end{figure*}

For the polygonal evaluation, our method compares with the three polygonal building extraction methods. As reported in Table \ref{tab:crowdai_poly}, the performance of the proposed HiT significantly increases compared with other methods. Especially, the N ratio of HiT is closer to 1, which has redundant vertices for FrameField (N ration=$1.13>1.0$) and insufficient vertices for PolyWorld (N ratio=$0.93<1.0$). The results illustrate that our method generates more accurate vertices than polygonal building extraction methods. Moreover, our method has achieved the highest C-IoU score (88.6\%), demonstrating that our method has better balanced building segmentation and geometric complexity among comparison methods. To measure the performance in polygonal building shape and structure, HiT obtains the lowest MTA value, which means the lower the MTA value, the better the performance. Specifically, HiT gets 31.7\% of the MTA indicator, which is 1.4\%, 1.8\%, and 1.2\% lower than PolyMapper, FrameField, and PolyWorld, respectively.

The results in terms of instance segmentation and polygonal metrics have proved that our model has a high ability for accurately building segmentation with a more precise polygonal structure. Besides, a comprehensive comparison has verified our proposed HiT's superiority and effectiveness.

\noindent \textbf{Qualitative Comparison}.
Figure \ref{fig:crowdaires} and \ref{fig:crowdaires2} shows some visualization results generated by our approach and the comparison methods for qualitative comparison. HiT can successfully extract all polygonal buildings with high quality, including buildings of different sizes, appearances, and shapes.

Compared with FrameField, HiT predicts more precise vertices regarding number and position. While FrameField can predict all buildings, it predicts many redundant vertices, which is not suitable for real-world applications. In addition, PolyWorld extracts polygonal buildings in a down-top pathway, resulting in error vertex detection or insufficient vertices. Although both HiT and PolyMapper represent building mapping as a vertex sequence prediction task, HiT can better handle occlusions due to predicting serialized vertices simultaneously by the polygon prediction head. Moreover, the designed hierarchical attention mechanism embeds geometric information into the building feature map so that HiT can deal with buildings under complex scenes. Besides, the introduced polygon head exploits the supervisions from vertex, edge, and polygon, leading to more robustness and generalization. The qualitative results further demonstrate the superiority of HiT.

\begin{table*} [htbp]
\centering
\caption{Results on the Inria Polygonized dataset under the instance segmentation metric. The best results are marked in bold.\label{tab:inria_inst}}
\begin{tabular}{lccccccc}
\hline
Method      & AP $\uparrow$ & $AP_{50}$ $\uparrow$ & $AP_{75}$ $\uparrow$ & AR $\uparrow$ & $AR_{50}$ $\uparrow$ & $AR_{75}$ $\uparrow$ & $F1$ $\uparrow$ \\
\hline
Mask RCNN \cite{he2017maskrcnn}       & 40.0 & 79.2 & 35.3 & 51.5 & 87.3 & 54.4 & 45.0 \\
PANet \cite{liu2018panet}             & 39.6 & 79.0 & 35.0 & 51.5 & 87.3 & 54.2 & 44.8 \\
PolyMapper \cite{li2019polymapper}    & 44.9 & 82.5 & 45.4 & 55.4 & 90.8 & 61.7 & 49.6 \\
FrameField \cite{girard2021ffl}       & 38.3 & 67.3 & 39.8 & 49.0 & 78.1 & 53.4 & 43.0 \\
HiT (ours)  & $\textbf{50.5}$ & $\textbf{86.1}$ & $\textbf{56.6}$ & $\textbf{60.6}$ & $\textbf{91.2}$ & $\textbf{71.0}$ & \textbf{55.1} \\
\hline
\end{tabular}
\end{table*}

\begin{table} [!tb]
\centering
\caption{Results on the Inria Polygonized dataset under the polygonal metric. ``N ratio (=1)'' denotes that the performance is better when the N ratio is closer to 1. The best results are marked in bold.\label{tab:inria_poly}}
\begin{tabular}{lccc}
\hline
Method       & C-IoU $\uparrow$ & MTA $\downarrow$ & N ratio (=1) \\
\hline
PolyMapper \cite{li2019polymapper}      & 41.5 & 34.4 & 1.6  \\
FrameField \cite{girard2021ffl}         & 49.4 & \textbf{32.4} & 2.1 \\
HiT (ours)  & $\textbf{64.5}$ & 33.2 & \textbf{0.8} \\
\hline
\end{tabular}
\end{table}

\begin{figure*}[htbp]
\centering
\includegraphics[width=\linewidth,scale=1.00]{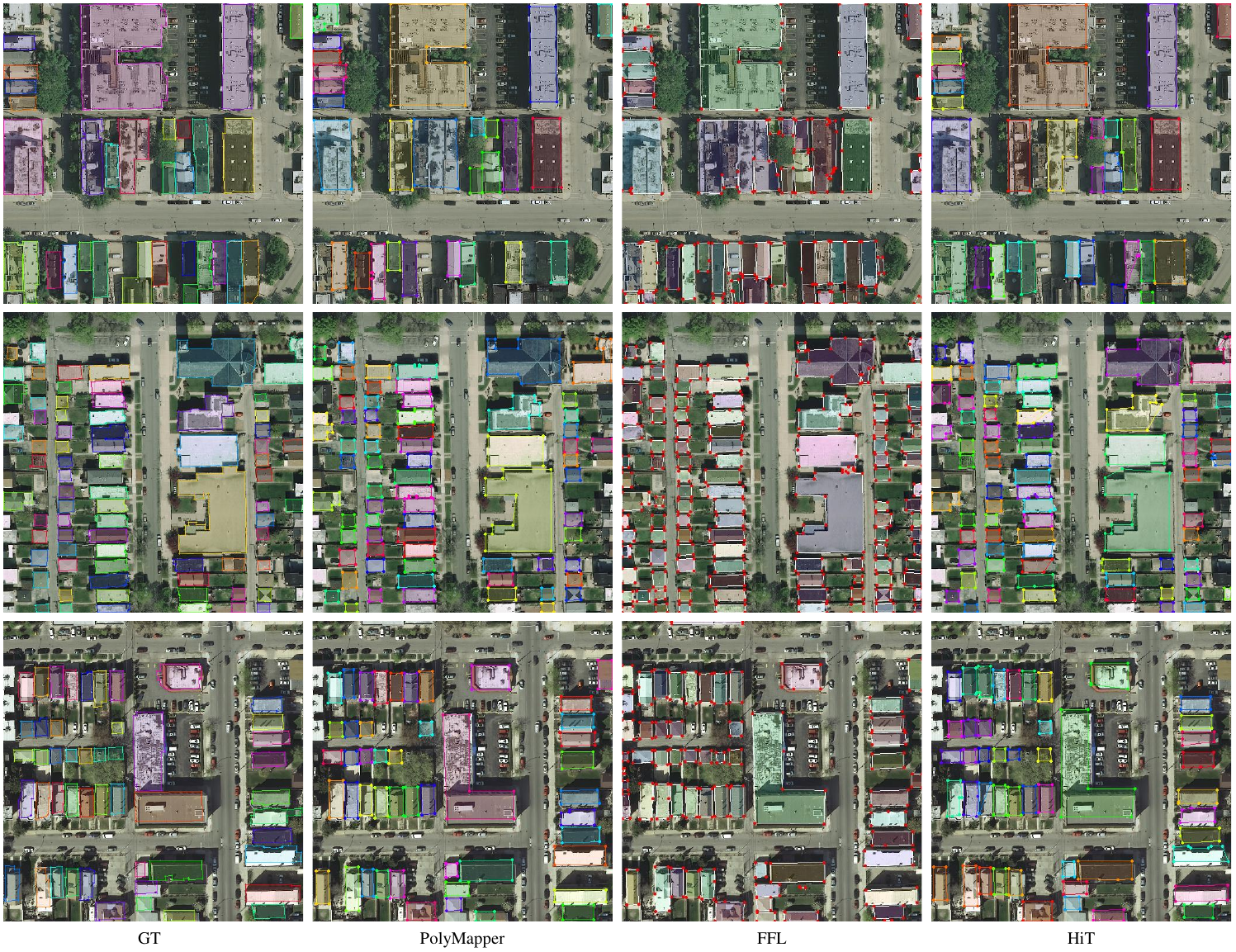}
\caption{Qualitative results on the Inria Polygonized dataset. The proposed HiT can better extract buildings in dense areas. Zoom in for a cleaner view.}
\label{fig:inriares}
\end{figure*}

\noindent \textit{(2) Results on Inria Polygonized dataset}.

In this subsection, we compare HiT on the Inria Polygonized dataset with Mask RCNN \cite{he2017maskrcnn}, PANet \cite{liu2018panet}, PolyMapper \cite{li2019polymapper}, and FrameField \cite{girard2021ffl} under the instance segmentation and polygonal metrics.

\noindent \textbf{Quantitative Evaluation}.
Quantitative results are reported in Table \ref{tab:inria_inst} and \ref{tab:inria_poly} from different methods under two metrics. For instance segmentation, our HiT has improved AP and AR on all the indicators shown in Table \ref{tab:inria_inst}. Compared with Mask RCNN, HiT achieves 50.5\% AP (+10.0\%), 60.6\% AR (+9.1\%), and 55.1\% F1 (+10.1\%), comprehensively indicating that HiT can generate highly accurate buildings. Since HiT is similar to Mask RCNN in the pipeline, the high performance of HiT demonstrates the effectiveness of the designed polygon prediction head. On the other hand, our HiT has improved the AP and AR scores by +5.6\% and +5.2\% compared with PolyMapper, respectively. In addition, the F1 score is +5.5\% higher than the SOTA polygonal building extraction method, as shown in Table \ref{tab:inria_inst}. Specifically, the $AP_{75}$ and $AR_{75}$ scores of the proposed HiT have outperformed the polygonal segmentation methods by +11.2\% and +9.3\%, respectively. These high improvements indicate that HiT generates building instances with high precision and recall.  

\begin{table} [tbp]
\centering
\caption{Model computational complexity. M and G denote million and gillion, respectively.}
\begin{tabular}{ccc}
\hline
Method & \#Params (M) $\downarrow$ & FLOPs (G) $\downarrow$ \\
\hline
Mask RCNN \cite{he2017maskrcnn}        & 43.8 & 114.7 \\
PANet \cite{liu2018panet}              & 47.7 & 123.1 \\
PolyMapper \cite{li2019polymapper}     & 53.8 & 717.6 \\
FrameField \cite{girard2021ffl}        & 76.7 & 204.3 \\
PolyWorld \cite{zorzi2022polyworld}    & 39.4 & 448.3 \\
HiT (ours)                             & 47.4 & 145.3 \\
\hline
\end{tabular}
\label{tab:modelcomplex}
\end{table}

Table \ref{tab:inria_poly} reports the polygonal evaluation from different polygonal building extraction methods. We can see that HiT significantly increases performances than comparison methods on all indicators. HiT detects buildings with more accurate vertices than other methods on the N ratio, but FrameField generates many more vertices with a large N ratio. Moreover, our method has significantly improved the C-IoU score by +15.1\%, illustrating that our method has better balanced building segmentation and geometric complexity.

The quantitative results on the Inria Polygonized dataset under different metrics have further proved that our model generates polygonal buildings with more accurate segmentation masks and precise polygonal structures.

\noindent \textbf{Qualitative Comparison}.
We show some qualitative results from our approach and comparison methods in Figure \ref{fig:inriares}. Compared with other methods, HiT extracts more accurate polygonal buildings of different sizes, appearances, and shapes. Moreover, HiT is more robust in dealing with images with complex scenes by using a transformer-based structure to predict the vertex sequence simultaneously. The visualization results consistently demonstrate the superiority of HiT.

\noindent \textit{(3) Discussion.}

\begin{figure}[htbp]
\centering
\includegraphics[width=\linewidth]{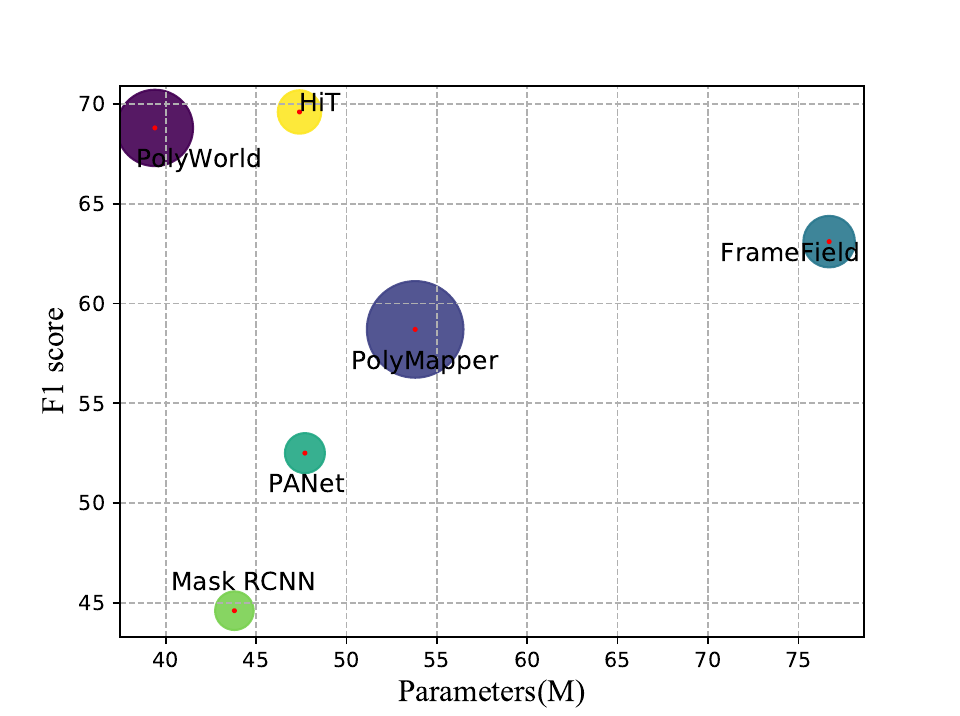}
\caption{Model complexity and F1 score comparison among different methods. The floating point operations (FLOPs (G)) is denoted by the radius of the circle.}
\label{fig:modelcmp}
\end{figure}

In this section, we evaluate model complexity among comparison methods and then discuss performance in terms of model structure, complexity, accuracy and robustness. For the model complexity, we calculate model parameters (\#Params (M)) and floating point operations (FLOPs (G)) by testing an image with a resolution of $512\times512$ on 1 GPU for all comparison methods. As reported in Table \ref{tab:modelcomplex}, our approach has lower \#Params and higher FLOPs than instance segmentation methods since HiT directly generates polygonal buildings rather raster building masks. Compared with polygonal building mapping, our model has much lower parameters, except PolyWorld \cite{zorzi2022polyworld}, and lower FLOPs than comparison methods, demonstrating our method can effectively extract polygonal buildings. Figure \ref{fig:modelcmp} shows comprehensive comparison between complexity and accuracy (F1 score). We use the radius of the circle to denote the floating point operations (FLOPs (G)). We can see that HiT can accurately extract polygonal buildings with lower \#Params and FLOPs. In addition, we discuss the model robustness using the CrowdAI dataset, which encompasses large-scale buildings across diverse regions, including urban, suburban, and rural landscapes. Quantitative results in Tables \ref{tab:crowdai_inst} and \ref{tab:crowdai_poly} reveal that HiT consistently achieves high performance and robustness compared to alternative methods. As depicted in Figures \ref{fig:crowdaires} and \ref{fig:crowdaires2}, HiT accurately delineates building polygons for sparse and dense buildings. HiT exhibits acceptable proficiency in handling occlusions or shadows, demonstrating its robustness.

HiT is built on a insightful perspective that a building polygon can be effectively formulated as a bidirectional vertex sequence. Hence, a simple polygon head is designed for serialized vertex prediction. The polygon head combine attention with convolution operations to encoding building features with rich geometric and semantic information in a hierarchical manner. Moreover, a bidirectional polygon loss guide the model to pay more attention on vertex positions and relationships, rather than the clockwise or counterclockwise orientation of the vertex sequence. Consequently, HiT has greater flexibility in polygonal building mapping.

\subsection{Ablation Study}
\label{sec:exp_ablation}
Our method designs a transformer-based polygon prediction head to extract building serialized vertices parallel with building classification and building bounding box regression by a two-stage detection framework. In the polygon prediction head, the encoder with the hierarchical attention operation is proposed to encode building feature maps with geometric and semantic information. The designed polygon prediction head is optimized using the serialized vertices prediction loss joint with the vertex and edge prediction loss. In this subsection, we perform ablation studies on the CrowdAI dataset to further analyze the effectiveness of the details of our approach, including the encoding mechanism, the decoding setting, and the training strategy. In the ablation experiments, we remove the encoder from the polygon prediction head and train the modified model on the small version of the CrowdAI dataset (CrowdAI-S), which is used as the baseline.

\begin{table*} [htbp]
\centering
\caption{Results for different encoding mechanisms in the polygon head on CrowdAI-S dataset under the instance segmentation metric. The best results are marked in bold.\label{tab:enc_ablation}}
\begin{tabular}{lccccccc}
\hline
Method      & AP $\uparrow$ & $AP_{50}$ $\uparrow$ & $AP_{75}$ $\uparrow$ & AR $\uparrow$ & $AR_{50}$ $\uparrow$ & $AR_{75}$ $\uparrow$ & $F1$ $\uparrow$ \\
\hline
Baseline                  & 31.1 & 64.1 & 27.5 & 44.3 & 78.4 & 45.9 & 36.5 \\
(a) Original              & 36.6 & 72.3 & 34.4 & 48.5 & 82.4 & 51.8 & 41.7 \\
(b) Vertex-enhanced       & 34.0 & 68.9 & 30.9 & 46.4 & 80.4 & 48.8 & 39.2 \\
(c) Edge-enhanced         & 31.7 & 65.5 & 27.9 & 44.7 & 78.9 & 46.3 & 37.1 \\
(d) Vertex-edge-enhanced  & 37.5 & 74.1 & 35.7 & 48.7 & 82.9 & 52.1 & 42.4 \\
(e) Vertex-wise           & 32.0 & 66.4 & 28.3 & 45.1 & 79.2 & 47.0 & 37.4 \\
(f) Edge-wise             & 33.7 & 68.3 & 30.6 & 46.4 & 80.2 & 48.9 & 39.0 \\
(g) Vertex-edge-wise      & 36.6 & 72.6 & 34.7 & 48.2 & 82.4 & 51.6 & 41.6 \\
(h) Hierarchical (ours)   & $\textbf{38.5}$ & $\textbf{75.3}$ & $\textbf{37.6}$ & $\textbf{49.3}$ & $\textbf{83.5}$ & $\textbf{53.2}$ & \textbf{43.2} \\
\hline
\end{tabular}
\end{table*}

\begin{figure}[!tp]
\centering
\includegraphics[width=\linewidth]{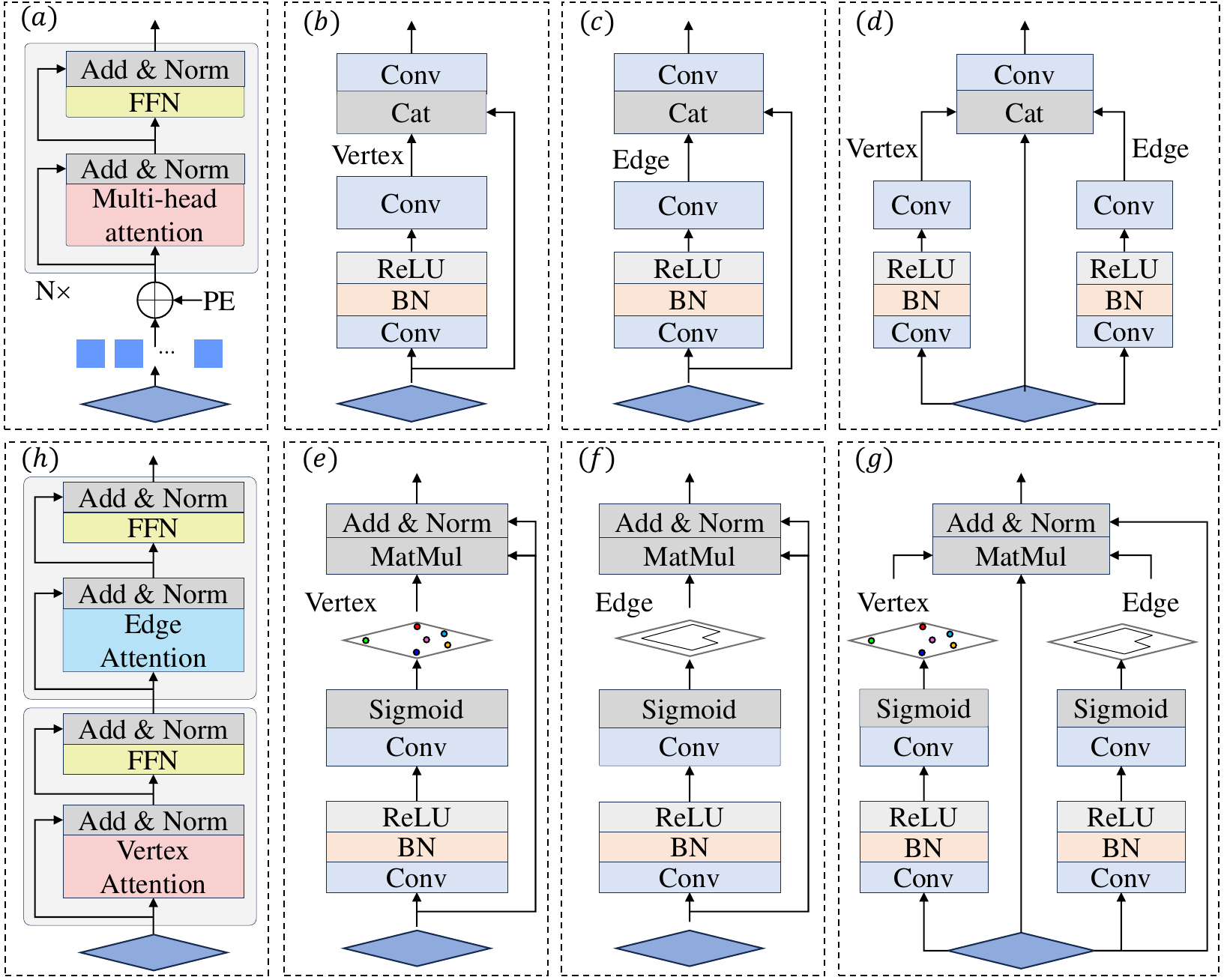}
\caption{Illustration of the vertex-level and the edge-level attention operations. The vertex-level and the edge-level attention replace the original self-attention mechanism to encode the building feature map, avoiding the complexity and speeding up the convergence speed by introducing the geometric information in terms of vertex and edge levels.}
\label{fig:enc_ablation}
\end{figure}

\noindent \textbf{(1) Encoding mechanism}.
The encoder of the polygon head plays a significant role in the serialized vertex prediction. In the encoding ablation experiments, we conduct eight different encoding mechanisms and remove the encoder from the polygon prediction head as a baseline. Figure \ref{fig:enc_ablation} shows different encoding methods discussed in the following.

\textbf{(a)} The original encoding mechanism. The original encoding manner uses an element-wise self-attention operation to obtain the relationship from the input sequence. As shown in Figure \ref{fig:enc_ablation}(a), the flattened building feature map is processed through N identical layers with self-attention and FFN operations.

\textbf{(b)} The vertex-enhanced mechanism. We use convolutional operations to get a vertex feature map shown in Figure \ref{fig:enc_ablation}(b). Then, the building and vertex feature maps are concatenated to enhance vertex information.

\textbf{(c)} The edge-enhanced mechanism. In Figure \ref{fig:enc_ablation}(c), the edge-enhanced encoding fashion is similar to the vertex-enhanced encoding, which can introduce edge information for serialized vertex prediction.

\textbf{(d)} The vertex-edge-enhanced mechanism. This encoding method obtains the vertex and edge features separately and then concatenates with the building feature, evaluating a joint enhanced encoding in vertex and edge levels shown in Figure \ref{fig:enc_ablation}(d).

\textbf{(e)} The vertex-wise attention mechanism. We formulate the building vertex probabilities as attention weights and multiply them with the building feature, enhancing vertex information in the building feature. Besides, the short-cut connection and layer normalization are used to get the building embedding, as described in Figure \ref{fig:enc_ablation}(e).

\textbf{(f)} The edge-wise attention mechanism. Like vertex-wise attention, the edge-wise attention mechanism multiplies edge prediction probabilities with the building feature, followed by short-cut connection and layer normalization operations.

\textbf{(g)} The vertex-edge-wise attention. We simultaneously exploit vertex-wise and edge-wise attention products to enhance the building feature map, as shown in Figure \ref{fig:enc_ablation}(g).

\textbf{(h)} The hierarchical attention mechanism. Motivated by the original encoding mode, the hierarchical attention mechanism replaces the self-attention operation with vertex-wise and edge-wise attentions due to the sparsity of serialized vertices in the building feature map B.

As shown in Table \ref{tab:enc_ablation}, the hierarchical attention mechanism significantly improves all the evaluation metrics compared with other encoding methods. We can see that encoding methods in concatenation and multiplication manners improve performance in all the indicators and show comparable performance to the original encoding method, proving that geometric information and the original encoding pipeline are effective in feature encoding. Motivated by this observation, we replace self-attention with hierarchical attention to introduce geometric information in building embeddings.

\noindent \textbf{(2) Decoding setting}.
The polygon head uses N identical decoder blocks to predict serialized vertices simultaneously. In each decoder block, the multi-head self-attention operation encodes the relationship among all the vertex queries. In this ablation experiment, we test head number \textit{H} and block number \textit{N} to select the optimal hyper-parameters.
Finally, we can observe from Table \ref{tab:dec_ablation} that optimal hyper-parameters are set as \textit{H}=4 and \textit{N}=8.

\begin{table} [htbp]
\centering
\caption{Results for different hyper-parameters in the decoder of the polygon head. \textit{H} and \textit{N} represent head number and block number. The best result is marked in bold.\label{tab:dec_ablation}}
\begin{tabular}{lcccccccc}
\hline
 & \multicolumn{2}{c}{H=1} & \multicolumn{2}{c}{H=2} & \multicolumn{2}{c}{H=4} & \multicolumn{2}{c}{H=8} \\
\cmidrule(r){2-3}  \cmidrule(r){4-5} \cmidrule(r){6-7} \cmidrule(r){8-9}
~ & $AP$ & $AP_{50}$ & $AP$ & $AP_{50}$ & $AP$ & $AP_{50}$ & $AP$ & $AP_{50}$  \\
\hline
N=1   & 31.1 & 65.5 & 33.4 & 68.6 & 35.3 & 72.3 & 35.8 & 72.5    \\
N=2   & 33.1 & 67.8 & 34.9 & 70.6 & 32.8 & 67.0 & 36.9 & 73.3    \\
N=4   & 31.4 & 65.9 & 31.9 & 67.6 & 34.0 & 68.3 & 33.4 & 67.8    \\
N=6   & 33.0 & 67.4 & 34.3 & 69.3 & 35.8 & 71.6 & 33.4 & 67.9    \\
N=8   & 34.4 & 70.4 & 37.1 & 73.8 & \textbf{38.5} & \textbf{75.3} & 37.3 & 73.2    \\
\hline
\end{tabular}
\end{table}

\noindent \textbf{(3) Training strategy}.
HiT is jointly trained by building classification, bounding box regression, and polygon prediction losses. In this ablation study, we select the optimal weighting coefficients $\lambda_{cls}$, $\lambda_{bbox}$, and $\lambda_{poly}$ to balance different modules. In Table \ref{tab:train_ablation}, the training objective is optimal when weighting coefficients are all set to 1.0.

\begin{table} [htbp]
\centering
\caption{Results for weighting coefficients $\lambda_{cls}$, $\lambda_{bbox}$, and $\lambda_{poly}$ selection in the joint training. The best result is marked in bold.\label{tab:train_ablation}}
\begin{tabular}{ccccccc}
\hline
$\lambda_{cls}$ & $\lambda_{bbox}$ & $\lambda_{poly}$ & $AP$ & $AP_{50}$ & $AR$ & $AR_{50}$ \\
\hline
1.0  & 1.0  & 0.01 & 31.9 & 64.6 & 44.6 & 76.0      \\
1.0  & 1.0  & 0.1  & 34.3 & 69.1 & 46.6 & 79.4      \\
10.0 & 10.0 & 0.1  & 32.1 & 65.3 & 44.7 & 76.0      \\
10.0 & 10.0 & 1.0  & 34.3 & 69.1 & 46.8 & 79.3      \\
1.0  & 1.0  & 1.0  & \textbf{38.5} & \textbf{75.3} & \textbf{49.3} & \textbf{83.5}      \\
\hline
\end{tabular}
\end{table}

\section{Conclusion}
We have presented HiT for automatically building mapping from remote sensing images. In this paper, we represent a building with serialized vertices, which can be formulated as a bidirectional vertex sequence. Based on this new observation, we apply a hierarchical transformer-based structure to predict serialized vertices. In the hierarchical transformer, we combine the CNN operation and transformer structure to embed semantic and geometry information, obtaining more effective building representations and capturing better building boundaries and corners. Moreover, we introduce a novel bidirectional polygon loss with bidirectional properties to train HiT end-to-endly. Finally, our extensive experiments illustrate that HiT significantly outperforms state-of-the-art methods, demonstrating its superiority.


\bibliographystyle{IEEEtran}
\bibliography{paper}

\ifCLASSOPTIONcaptionsoff
  \newpage
\fi

\begin{IEEEbiography}
[{\includegraphics[width=1in,height=1.25in,clip,keepaspectratio]{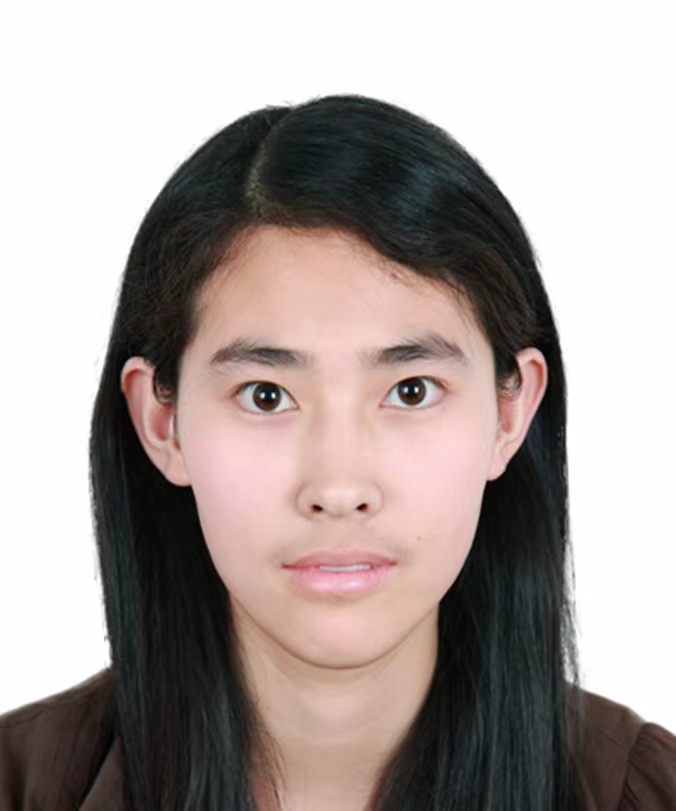}}]
{Mingming Zhang}
received the BS degree in information and computer science from Liaoning University, Shenyang, China and the MS degree in software engineering from Beihang University, Beijing, China. She is currently pursuing the Ph.D. degree in computer science with the Laboratory of Intelligent Recognition and Image Processing, School of Computer Science and Engineering, Beihang University. Her research interests include remote sensing image analysis and computer vision.
\end{IEEEbiography}

\begin{IEEEbiography}
[{\includegraphics[width=1in,height=1.25in,clip,keepaspectratio]{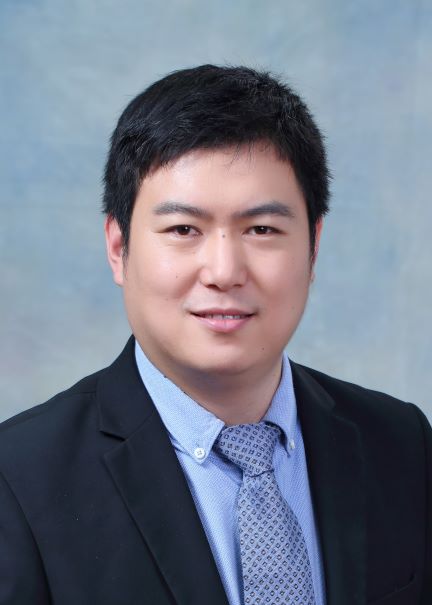}}]
{Qingjie Liu}
received the BS degree in computer science from Hunan University, Changsha, China, in 2007, and the Ph.D. degree in computer science from Beihang University, Beijing, China, in 2014. He is currently an Associate Professor with the School of Computer Science and Engineering, Beihang University. He is also a Distinguished Research Fellow with the Hangzhou Institute of Innovation, Beihang University, Hangzhou. His current research interests include image fusion, object detection, image segmentation, and change detection. He is a member of the IEEE.
\end{IEEEbiography}

\begin{IEEEbiography}
[{\includegraphics[width=1in,height=1.25in,clip,keepaspectratio]{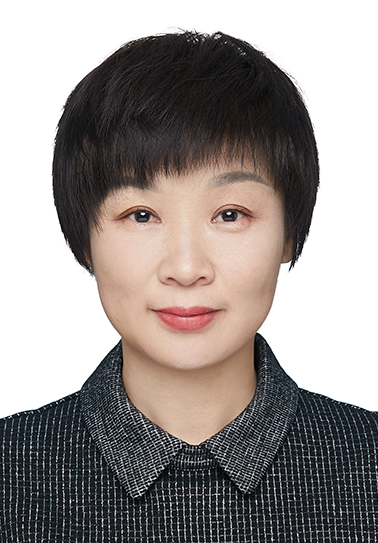}}]
{Yunhong Wang}
received the BS degree in electronic engineering from Northwestern Polytechnical University, Xi’an, China, in 1989, and the MS and Ph.D. degrees in electronic engineering from the Nanjing University of Science and Technology, Nanjing, China, in 1995 and 1998, respectively. 

She was with the National Laboratory of Pattern Recognition, Institute of Automation, Chinese Academy of Sciences, Beijing, China, from 1998 to 2004. Since 2004, she has been a Professor with the School of Computer Science and Engineering, Beihang University, Beijing, where she is also the Director of the Laboratory of Intelligent Recognition and Image Processing. Her research interests include biometrics, pattern recognition, computer vision, data fusion, and image processing. She is a Fellow of IEEE, IAPR, and CCF.
\end{IEEEbiography}

\end{document}